\documentclass[letterpaper, 10 pt, journal, twoside]{IEEEtran}

\usepackage[T1]{fontenc}
\usepackage{times}
\usepackage{amsmath}
\usepackage{amssymb}
\usepackage{graphicx}
\usepackage[ruled,linesnumbered]{algorithm2e}
\usepackage{booktabs}
\usepackage{float}
\usepackage{bigstrut}
\usepackage{multirow}
\usepackage{balance}
\usepackage[hidelinks=true,bookmarks=False,colorlinks=true]{hyperref}
\usepackage{lineno}
\usepackage{amsopn}
\usepackage{comment}
\usepackage{color}
\usepackage{cite}
\usepackage{algorithmic}
\usepackage{tabularx}
\usepackage{xcolor}
\usepackage{mathrsfs}
\usepackage{makecell}

\makeatletter

\newcommand{\Rmnum}[1]{\expandafter\@slowromancap\romannumeral #1@}
\makeatother

\newcolumntype{L}[1]{>{\raggedright\arraybackslash}p{#1}}
\newcolumntype{C}[1]{>{\centering\arraybackslash}p{#1}}
\newcolumntype{R}[1]{>{\raggedleft\arraybackslash}p{#1}}

\usepackage{pifont}
\newcommand{\cmark}{\ding{51}}%

\def\ie{\textit{i.e.}}
\def\eg{\textit{e.g.}}
\def\etal{\textit{et al.}}

\title{PVStereo: Pyramid Voting Module for End-to-End Self-Supervised Stereo Matching}

\author{Hengli Wang, Rui Fan, Peide Cai, and Ming Liu, \IEEEmembership{Senior Member, IEEE}
\thanks{\textit{(Corresponding author: Ming Liu.)}}
\thanks{H. Wang, P. Cai and M. Liu are with the Department of Electronic and Computer Engineering, the Hong Kong University of Science and Technology, Clear Water Bay, Kowloon, Hong Kong SAR, China (email: hwangdf@connect.ust.hk; pcaiaa@connect.ust.hk; eelium@ust.hk).}
\thanks{R. Fan is with both the Department of Computer Science and Engineering and the Department of Ophthalmology, the University of California San Diego, La  Jolla, CA 92093, U.S. (email: rui.fan@ieee.org).} 
\thanks{H. Wang and R. Fan contributed equally to this work.}
}

\begin{document}

\maketitle

\begin{abstract}
    Supervised learning with deep convolutional neural networks (DCNNs) has seen huge adoption in stereo matching. However, the acquisition of large-scale datasets with well-labeled ground truth is cumbersome and labor-intensive, making supervised learning-based approaches often hard to implement in practice. To overcome this drawback, we propose a robust and effective self-supervised stereo matching approach, consisting of a pyramid voting module (PVM) and a novel DCNN architecture, referred to as OptStereo. Specifically, our OptStereo first builds multi-scale cost volumes, and then adopts a recurrent unit to iteratively update disparity estimations at high resolution; while our PVM can generate reliable semi-dense disparity images, which can be employed to supervise OptStereo training. Furthermore, we publish the HKUST-Drive dataset, a large-scale synthetic stereo dataset, collected under different illumination and weather conditions for research purposes. Extensive experimental results demonstrate the effectiveness and efficiency of our self-supervised stereo matching approach on the KITTI Stereo benchmarks and our HKUST-Drive dataset. PVStereo, our best-performing implementation, greatly outperforms all other state-of-the-art self-supervised stereo matching approaches. Our project page is available at \url{sites.google.com/view/pvstereo}.
\end{abstract}


\section{Introduction}
\label{sec.intro}
\IEEEPARstart{H}{umans} live in a three-dimensional (3D) world, but our eyes can only perceive objects in two dimensions. The miracle of human depth perception is due to our brain's ability to analyze the difference between the two two-dimensional (2D) images which are projected on the retinas of our eyes. In a broad sense, each pair of corresponding points on the retinas send signals to the binocular neurons in the primary visual cortex, which then estimates the relative positional difference between each pair of correspondence points \cite{qian1997binocular}. This relative positional difference is generally referred to as \textit{disparity} \cite{fan2020computer}.

Similarly, two synchronized digital cameras can be utilized to extrapolate the 3D information of a given scenario. This process is typically known as \textit{stereo vision} or \textit{stereo matching} \cite{fan2020computer}. Stereo vision is a critical technology employed in many robotics and computer vision applications, such as freespace segmentation \cite{fan2020sne,wang2020applying,fan2021learning} and anomaly detection \cite{wang2019self,fan2020we,wang2021dynamic}. Existing stereo matching approaches are either mathematical modeling-based or data-driven ones. The former ones generally formulate stereo matching as block matching or energy minimization problems \cite{dinh2019disparity}, while the latter ones typically employ data-driven classification and/or regression models, \eg, convolutional neural networks (CNNs), to learn a feasible solution to stereo matching. With recent advances in deep learning, many researchers have resorted to deep CNNs (DCNNs) for stereo matching \cite{psmnet,gwcnet,zhang2020adaptive}. However, these approaches generally require a large amount of human-annotated training data to learn the best DCNN parameters. Such a data labeling process can be extremely time-consuming and labor-intensive. Furthermore, the limitation in DCNN generalization often fails these approaches when adapting to new scenarios in practice. Hence, there is a strong motivation to develop a self-supervised stereo matching approach, which does not require any human-annotated disparity ground truth to learn the best DCNN parameters \cite{smolyanskiy2018importance}.

\begin{figure}[t]
    \centering
    \includegraphics[width=0.99\linewidth]{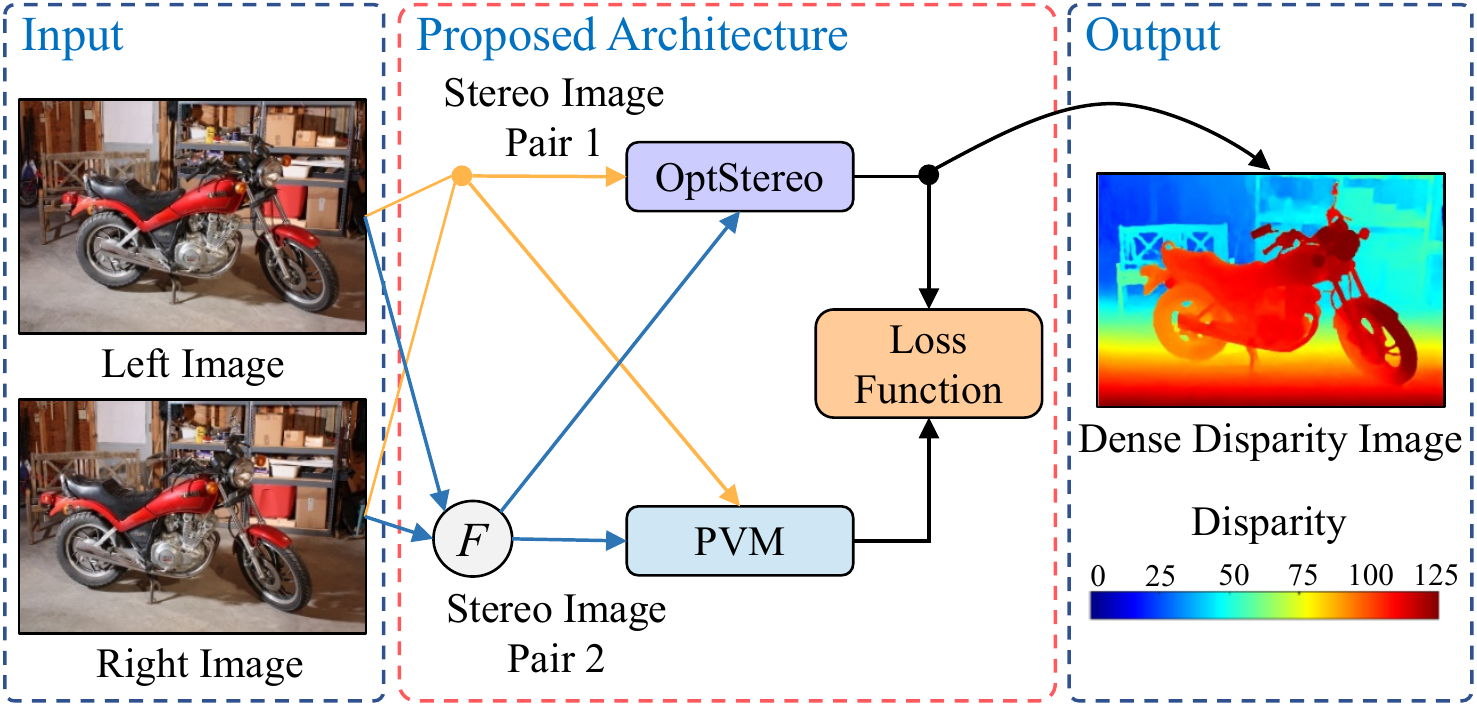}
    \caption{A schema of our proposed self-supervised stereo matching approach, where OptStereo is combined with PVM for self-supervised disparity estimation, and $F$ refers to a left-to-right flipping operation.}
    \label{fig.framework}
\end{figure}

Hence in this paper, we propose a novel approach for self-supervised stereo matching, as illustrated in Fig. \ref{fig.framework}. Specifically, we develop a module named \textit{Pyramid Voting Module} (PVM), which can be deployed in any supervised stereo matching DCNN, converting it into a self-supervised approach. With the use of our PVM, researchers will no longer require any hand-labeled data to train DCNNs for dense stereo matching, which greatly alleviates the labor for disparity ground truth labeling. Moreover, existing stereo matching DCNNs mainly rely on 3D convolutions \cite{psmnet,gwcnet,zhang2020adaptive} or  coarse-to-fine paradigms \cite{tankovich2020hitnet,wang2019anytime,yee2020fast}. Unfortunately, the former ones can consume a lot of computational resources, resulting in their limited ability to achieve real-time performance; while the latter ones generally suffer from the accumulated errors from the coarse pyramid level, which can in turn propagate to subsequent levels and further cause significant performance degradation. To address these issues, we propose a new DCNN architecture, referred to as \textit{OptStereo}, which is motivated by traditional optimization-based approaches. Our OptStereo first builds multi-scale cost volumes, and then adopts a recurrent unit to iteratively update disparity estimations at high resolution. This novel architecture enables our OptStereo to 1) avoid the error accumulation problem in coarse-to-fine paradigms and 2) achieve a great trade-off between accuracy and efficiency. Furthermore, we publish the \textit{HKUST-Drive dataset}, a large-scale synthetic stereo dataset created under different illumination and weather conditions, available at \url{sites.google.com/view/pvstereo} for research purposes. It contains 11568 pairs of stereo driving scene images and the corresponding dense ground-truth disparity images. To validate the effectiveness and efficiency of our proposed self-supervised stereo matching approach, we conduct extensive experiments on the popular KITTI Stereo benchmarks \cite{kitti12,kitti15} as well as our HKUST-Drive dataset. Extensive experimental results demonstrate that our best-performing implementation, \textit{PVStereo}, outperforms all other self-supervised stereo matching approaches. The major contributions of this paper can be summarized as follows:
\begin{itemize}
    \item PVM, a novel module capable of generating reliable semi-dense disparity images that can be used for supervising DCNN training.
    \item OptStereo, a novel DCNN architecture that can achieve a great trade-off between accuracy and efficiency for stereo matching.
    \item HKUST-Drive, a large-scale synthetic stereo dataset collected under different illumination and weather conditions for research purposes.
\end{itemize}

The remainder of this paper is organized as follows: Section~\ref{sec.related_work} introduces existing traditional and data-driven approaches for stereo matching. Then, Section~\ref{sec.methodology} presents our proposed framework for self-supervised stereo matching. The experimental results are illustrated in Section~\ref{sec.experiments}. Finally, Section~\ref{sec.conclusions} summarizes the paper.

\section{Related Work}
\label{sec.related_work}

\subsection{Traditional Stereo Matching Approaches}
Traditional stereo matching approaches can be classified into three main categories: 1) local, 2) global, and 3) semi-global \cite{dinh2019disparity}. Local algorithms simply select a group of image blocks from the target image and match them with a fixed image block selected from the reference image \cite{fan2018road, hirschmuller2008evaluation}.
The desirable disparities possess either the lowest matching costs, \eg, sum of absolute differences (SAD), or the highest correlation costs, \eg, normalized cross-correlation (NCC) \cite{tippetts2016review}.  The optimization strategy utilized in local algorithms is typically referred to as winner-takes-all (WTA) \cite{fan2018real}.

Unlike local algorithms, global algorithms typically formulate stereo matching as an energy minimization problem, which can be solved by some Markov random field (MRF)-based optimization approaches, such as graph cuts (GC) \cite{boykov2001fast} and belief propagation (BP) \cite{ihler2005loopy}. Semi-global matching (SGM) \cite{lee2017memory} approximates the MRF inference by performing cost aggregation along all directions in the image to improve both the accuracy and efficiency for stereo matching \cite{fan2020rethinking}. However, traditional methods are either inaccurate (local algorithms) or computationally intensive (global algorithms). With recent advances in deep learning, data-driven approaches can achieve a great trade-off between accuracy and efficiency.

\subsection{Data-Driven Stereo Matching Approaches}
\subsubsection{Supervised Stereo Matching Approaches}
Supervised stereo matching approaches can be classified into three categories: 1) learning better feature correspondences, 2) learning better regularization, and 3) learning dense disparity images in an end-to-end way. The first category of approaches utilize the learned distinguishable features to compute stereo matching costs, and then apply traditional cost aggregation and regularization for disparity estimation \cite{vzbontar2016stereo}. The second category of approaches learn both regularization and cost aggregation, \eg, the spatial-variant penalty-parameters in SGM \cite{seki2017sgm}.

Recently, researchers have turned their focuses towards the third category, due to its excellent performance on public benchmarks. Such end-to-end approaches generally rely on 3D convolutions \cite{psmnet,gwcnet,zhang2020adaptive} or coarse-to-fine paradigms \cite{tankovich2020hitnet,wang2019anytime,yee2020fast}. Specifically, Chang \etal \cite{psmnet} proposed PSMNet, a pyramid stereo matching network consisting of spatial pyramid pooling and several 3D convolutional layers. GwcNet \cite{gwcnet} and AcfNet \cite{zhang2020adaptive} were developed based on PSMNet for further performance improvement. However, 3D convolutions can consume a lot of computational resources, making these approaches difficult to perform in practice. To improve the DCNN inference speed, some researchers have adopted coarse-to-fine paradigms to replace 3D convolutions. Specifically, Tankovich \etal \cite{tankovich2020hitnet} proposed HITNet, which generates disparity predictions hierarchically from 1/64 resolution to 1/4 resolution. Similarly, Wang \etal \cite{wang2019anytime} and Yee \etal \cite{yee2020fast} also followed this paradigm to improve the DCNN inference speed. However, these approaches typically have limited capability to recover errors from coarse resolutions, which can lead to significant performance degradation. Unlike the above-mentioned prior works, our proposed OptStereo employs a recurrent unit to iteratively updates disparity estimations at high resolution, which helps achieve a great trade-off between accuracy and efficiency for stereo matching.

\begin{figure}[t]
    \centering
    \includegraphics[width=0.99\linewidth]{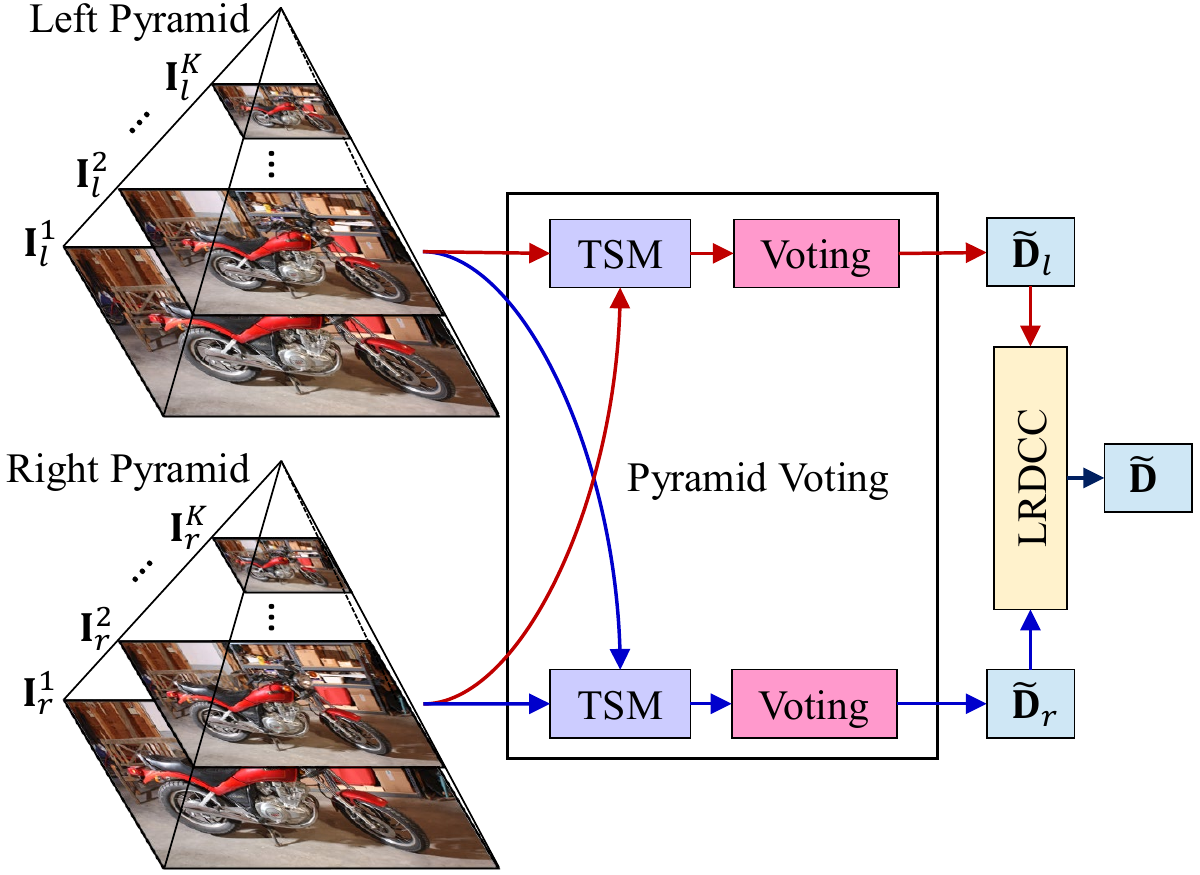}
    \caption{An illustration of our proposed PVM. The input stereo images are processed to generate reliable semi-dense disparity images for supervising DCNN training. The red and blue paths are used to produce the left and right semi-dense disparity images, respectively.}
    \label{fig.pvm}
\end{figure}

\subsubsection{Unsupervised Stereo Matching Approaches}
To reduce the labor for disparity ground truth labeling, many researchers have proposed unsupervised stereo matching approaches \cite{zhong2017self, tulyakov2017weakly, flow2stereo}. Specifically, Zhong \etal \cite{zhong2017self} proposed a self-supervised stereo matching approach, which uses a novel training loss to exploit the loop constraint in image warping process and handle the texture-less areas. Given coarse information about the scenes and the optical system, Tulyakov \etal \cite{tulyakov2017weakly} developed an approach to generate disparity estimations in a weakly-supervised manner. Moreover, Liu \etal \cite{flow2stereo} proposed Flow2Stereo, which leverages the geometric constraints behind stereoscopic videos to perform disparity and optical flow estimation in a self-supervised manner. Different from these approaches, we propose PVM in this paper for reliable semi-dense disparity generation. The generated disparity images are then used to supervise DCNN training. Compared to the prior works, our PVM is easy-to-use, efficient and accurate. Extensive experimental results provided in Section \ref{sec.experiments} demonstrate the superiority of our PVStereo over all other state-of-the-art self-supervised stereo matching approaches.

\section{Methodology}
\label{sec.methodology}

\subsection{Pyramid Voting Module}
Our PVM can produce a reliable semi-dense disparity image $\tilde{\mathbf{D}}$ under a multi-scale disparity voting strategy, as illustrated in Fig. \ref{fig.pvm}. The produced $\tilde{\mathbf{D}}$ can be utilized to supervise DCNNs in learning dense disparity estimation, and therefore, our PVM can convert any supervised stereo matching DCNN into a self-supervised approach.

Our PVM is designed based on two hypotheses:
\begin{enumerate}
    \item confident disparities possess similar values, and
    \item their matching costs or correlations are consistent,
\end{enumerate}
regardless of image resolution. Therefore, our PVM aims at seeking out consistent disparities among multi-scale stereo image pairs from two pyramids. Given a pair of left and right stereo images $\mathbf{I}_{l}$ and $\mathbf{I}_{r}$, PVM first generates two groups of stereo image pairs, constructing a left and a right pyramid, respectively. One group is used to produce the left semi-dense disparity image $\tilde{\mathbf{D}}_l$ (see the red flow in Fig. \ref{fig.pvm}), while the other one is used to produce the right semi-dense disparity image $\tilde{\mathbf{D}}_r$ (see the blue flow in Fig. \ref{fig.pvm}). Each group contains a collection of $K$ stereo image pairs at different scales, as illustrated as $(\mathbf{I}^{1}_{l}, \mathbf{I}^{1}_{r}), \dots, (\mathbf{I}^{K}_{l}, \mathbf{I}^{K}_{r})$. In this paper, $\mathbf{I}^{k}_{l,r}$  represents an $\mathbf{I}_{l,r}$ downsampled by a scale $k+\epsilon$, where $\epsilon\in(-1,1)$ is a random scalar and $\epsilon=0$ when $k=1$. Each generated stereo image pair $(\mathbf{I}^{k}_{l}, \mathbf{I}^{k}_{r})$ can separately produce a left and a right disparity image $\tilde{\mathbf{D}}^{k}_{l,r}$ via a traditional stereo matching algorithm (abbreviated as TSM in Fig. \ref{fig.pvm})  \cite{hirschmuller2008evaluation}. Based on the above hypotheses, a representation $\mathbf{C}$ can be obtained:
\begin{equation}
    \begin{split}
        \mathbf{C}(\mathbf{p}) = & \Bigg(\sqrt{\frac{1}{K}\sum_{k=1}^{K}(\tilde{\mathbf{D}}^{k}_{l,r}(\mathbf{p})-\frac{1}{K}\sum_{k=1}^{K}\tilde{\mathbf{D}}^{k}_{l,r}(\mathbf{p}))^2},
        \\
        & \sqrt{\frac{1}{K}\sum_{k=1}^{K}(\tilde{\mathbf{c}}^{k}_{l,r}(\mathbf{p})-\frac{1}{K}\sum_{k=1}^{K}\tilde{\mathbf{c}}^{k}_{l,r}(\mathbf{p}))^2}
        \Bigg)
        ,
        \label{eq.sdc}
    \end{split}
\end{equation}
where $\mathbf{p}$ is an image pixel; and $\tilde{\mathbf{c}}^{k}_{l,r}\in[0,1]$ denotes the normalized inverse stereo matching cost or normalized correlation (a better stereo matching corresponds to a higher $\tilde{\mathbf{c}}^{k}_{l,r}$ value). A voting map $\mathbf{V}$ can then be obtained, where $\mathbf{V}(\mathbf{p})=\delta(\mathbf{C}(\mathbf{p},1),\kappa_1)+\delta(\mathbf{C}(\mathbf{p},2),\kappa_2)$. $\kappa_1$ and $\kappa_2$ are two thresholds; and $\delta(x,y)=0$ when $x<y$, otherwise, $\delta(x,y)=1$. Generating a denser $\tilde{\mathbf{D}}_{l,r}$ requires higher $\kappa_1$ and $\kappa_2$. By finding the disparities at which $\mathbf{V}(\mathbf{p})=0$, a reliable semi-dense confident disparity image $\tilde{\mathbf{D}}_{l,r}$ is produced. Finally, $\tilde{\mathbf{D}}_l$ and $\tilde{\mathbf{D}}_r$ are processed by a left-right disparity consistency check (LRDCC) operator to produce $\tilde{\mathbf{D}}$, which is further employed to supervise DCNN training. More details on the DCNN architecture and training phase will be discussed in the next subsections. 

\begin{figure*}[t]
    \centering
    \includegraphics[width=0.99\textwidth]{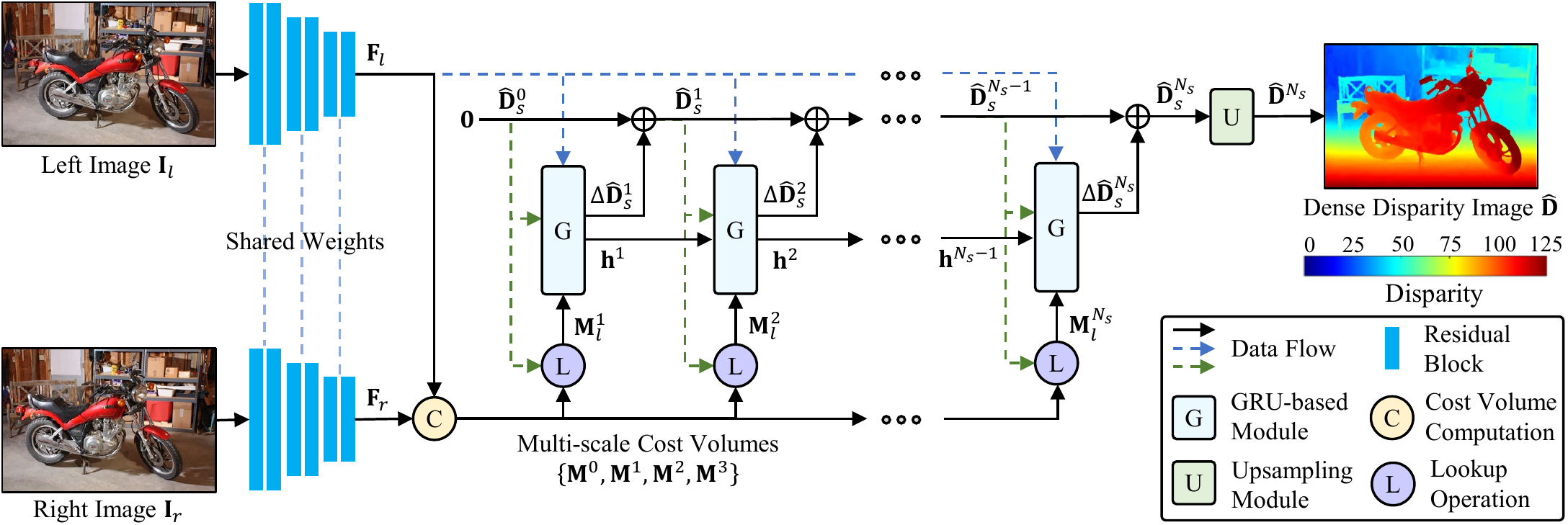}
    \caption{An illustration of our proposed OptStereo, which first builds multi-scale cost volumes, and then adopts a recurrent unit to iteratively update disparity estimations at high resolution.}
    \label{fig.optstereo}
\end{figure*}

\subsection{OptStereo}
Given a pair of left and right stereo images $\mathbf{I}_l$ and $\mathbf{I}_r$, our OptStereo is designed to estimate a dense disparity image $\hat{\mathbf{D}}$. Fig. \ref{fig.optstereo} illustrates the overview of our OptStereo, which consists of three stages, \ie, feature extraction, cost volume computation and iterative refinement.

\subsubsection{Feature Extraction} We use two residual networks \cite{he2016deep} that share weights to extract visual features $\mathbf{F}_l$ and $\mathbf{F}_r$ with the size of $H \times W \times C$ from $\mathbf{I}_l$ and $\mathbf{I}_r$, respectively. The spatial size of $\mathbf{F}_l$ and $\mathbf{F}_r$ is 1/8 of the input image resolution, and $C$ is set to be 256. Moreover, the residual networks includes six residual blocks, two at 1/2 resolution, two at 1/4 resolution and two at 1/8 resolution.

\subsubsection{Cost Volume Computation} In this stage, we compute visual similarity for all possible matching pairs between $\mathbf{F}_l$ and $\mathbf{F}_r$. Specifically, we construct a cost volume $\mathbf{M}^0 \in \mathbb{R}^{H \times W \times W} $ by computing the dot product between all possible matching pairs of feature vectors, which can be formulated as follows:
\begin{equation}
    \mathbf{M}^0(i,j,k) = \sum_{h=0}^{C} \mathbf{F}_l(i,j,h) \cdot \mathbf{F}_r(i,k,h).
\end{equation}
Following \cite{teed2020raft}, we further construct multi-scale cost volumes $\mathbf{M}^1$, $\mathbf{M}^2$ and $\mathbf{M}^3$ by employing average pooling on the last dimension of $\mathbf{M}^0$ respectively, where $\mathbf{M}^k$ has the size of $H \times W \times (W/2^k)$ for $k = 0, 1, 2, 3$. The multi-scale cost volumes $\{\mathbf{M}^0, \mathbf{M}^1, \mathbf{M}^2, \mathbf{M}^3\}$ store information about both large and small disparities, which can be used effectively to update disparity estimations in the iterative refinement stage.

To better utilize the multi-scale cost volumes, we introduce a lookup operation, which can extract values from $\{\mathbf{M}^0, \mathbf{M}^1, \mathbf{M}^2, \mathbf{M}^3\}$ to construct a local cost volume $\mathbf{M}_{l}$. Specifically, a dense disparity estimation $\hat{\mathbf{D}}_s$ with the size of $H \times W$ can map a point $\mathbf{p}=(i,j)$ in $\mathbf{F}_l$ to its correspondence $\mathbf{p}'=(i',j')=(i, j-\hat{\mathbf{D}}_{s}(\mathbf{p}))$ in $\mathbf{F}_r$. We also define a neighbor area around $j'$ as follows:
\begin{equation}
    U(j')_d = \left\{j'+\Delta j | \Delta j \in \mathbb{Z}, |\Delta j| \leq d \right\},
\end{equation}
where $d$ is the constant lookup distance. Then, we can extract values from $\mathbf{M}^k$ by adopting $(i,j)$ as the first two dimensional indexes and every element of $U(j'/2^k)_d$ as the last dimensional index for $k = 0, 1, 2, 3$. Please note that we employ bilinear sampling, since the adopted indexes are real numbers. In addition, we set $d = 4$, which corresponds to a search range of 256 pixels at the input image resolution. The values extracted from the multi-scale cost volumes are then concatenated into a local cost volume $\mathbf{M}_{l} \in \mathbb{R}^{H \times W \times 36}$, which provides useful visual similarity information around the possible matching pairs indicated by the dense disparity estimation for further refinement.

\subsubsection{Iterative Refinement} In this stage, we employ a GRU-based module \cite{cho2014learning} to iteratively update a sequence of dense disparity estimations $\{ \hat{\mathbf{D}}^1_s, \dots, \hat{\mathbf{D}}^{N_s}_s \}$ at the 1/8 resolution with an initialization $\hat{\mathbf{D}}^0_s = 0$. Specifically, in iteration $k$, we first use the above-mentioned lookup operation to compute the local cost volume $\mathbf{M}^{k}_{l}$ based on the previous dense disparity estimation $\hat{\mathbf{D}}^{k-1}_s$. Then, we denote the concatenation of $\hat{\mathbf{D}}^{k-1}_s$, $\mathbf{M}^{k}_{l}$ and $\mathbf{F}_l$ as $\mathbf{x}^k$, and send $\mathbf{x}^k$ to the GRU-based module, which has the following formulation:
\begin{equation}
    \begin{aligned}
        \mathbf{z}^{k}         & =\sigma\left(\operatorname{Conv}_{3 \times 3}\left(\left[\mathbf{h}^{k-1}, \mathbf{x}^{k}\right]\right)\right),                      \\
        \mathbf{r}^{k}         & =\sigma\left(\operatorname{Conv}_{3 \times 3}\left(\left[\mathbf{h}^{k-1}, \mathbf{x}^{k}\right]\right)\right),                      \\
        \tilde{\mathbf{h}}^{k} & =\tanh \left(\operatorname{Conv}_{3 \times 3}\left(\left[\mathbf{r}^{k} \odot \mathbf{h}^{k-1}, \mathbf{x}^{k}\right]\right)\right), \\
        \mathbf{h}^{k}         & =\left(1-\mathbf{z}^{k}\right) \odot \mathbf{h}^{k-1}+\mathbf{z}^{k} \odot \tilde{\mathbf{h}}^{k},
    \end{aligned}
    \label{eq.gru}
\end{equation}
where $\left[\cdot,\cdot\right]$, $\sigma$ and $\odot$ denote concatenation, sigmoid function and element-wise multiplication, respectively. The outputted hidden state $\mathbf{h}^{k}$ is then processed by two convolutional layers to generate the disparity update $\Delta \hat{\mathbf{D}}^{k}_s$, and the dense disparity estimation $\hat{\mathbf{D}}^{k}_s$ is updated by $\hat{\mathbf{D}}^{k}_s = \hat{\mathbf{D}}^{k-1}_s + \Delta \hat{\mathbf{D}}^{k}_s$. This process iterates until $N_s = 8$ is reached. Note that $\{ \hat{\mathbf{D}}^1_s, \dots, \hat{\mathbf{D}}^{N_s}_s \}$ are at the 1/8 resolution. We then employ an upsampling module that consists of an upsampling layer followed by two convolutional layers to generate the full resolution disparity estimations $\{ \hat{\mathbf{D}}^1, \dots, \hat{\mathbf{D}}^{N_s} \}$. In the inference phase, we take $\hat{\mathbf{D}}^{N_s}$ as the estimated dense disparity image $\hat{\mathbf{D}}$.

The architecture of our OptStereo is inspired by traditional optimization-based approaches. Specifically, the adoption of the GRU-based module mimics the updates of a first-order descent algorithm, and the bounded activations used in (\ref{eq.gru}) also encourage convergence to a fixed point \cite{teed2020raft}. Moreover, since our OptStereo iteratively updates disparity estimations at high resolution, it does not suffer from the error accumulation problem in the coarse-to-fine paradigm. Our OptStereo can also greatly minimize the trade-off between accuracy and efficiency for stereo matching due to its simple but effective architecture.

\subsection{Loss Function and Data Augmentation}
In the training phase, we optimize the parameters of our OptStereo by minimizing a loss function $\mathcal{L}$, which consists of three terms:
\begin{equation}
    \mathcal{L}=\mathcal{L}_{P}+\lambda_{1} \cdot \mathcal{L}_{R}+\lambda_{2} \cdot \mathcal{L}_{S},
    \label{eq.whole_loss}
\end{equation}
where $\mathcal{L}_{P}$ denotes the PVM guiding loss; $\mathcal{L}_{R}$ denotes the reconstruction loss; $\mathcal{L}_{S}$ denotes the smoothing loss; and $\lambda_{1}$ and $\lambda_{2}$ are two hyper-parameters to weight the contributions of the three above-mentioned loss terms.

The $\mathcal{L}_{P}$ term takes the reliable semi-dense disparity images generated by our PVM as the supervision for training. Specifically, we adopt the Huber loss function $l(\cdot)$ for the $\mathcal{L}_{P}$ term, which is defined as:
\begin{equation}
    \begin{gathered}
        \mathcal{L}_{P} = \frac{1}{\sum_{i=1}^{N_s} \gamma^{N_s-i}} \sum_{i=1}^{N_s} \frac{\gamma^{N_s-i}}{N_{\tilde{\mathbf{D}}}} \sum_{\mathbf{p}\in\tilde{\mathbf{D}}} l(|\tilde{\mathbf{D}}(\mathbf{p})-\hat{\mathbf{D}}^{i}(\mathbf{p})|), \\
        l(x) =\left\{\begin{array}{ll}
            {x-0.5,}     & {x \geq 1} \\
            {x^{2} / 2,} & {x<1}
        \end{array},\right.
    \end{gathered}
    \label{eq.guiding_loss}
\end{equation}
where $\tilde{\mathbf{D}}$ denotes the reliable semi-dense disparity image generated by our PVM; $\hat{\mathbf{D}}^{i}$ denotes the dense disparity image estimated by our OptStereo; $N_{\tilde{\mathbf{D}}}$ is the number of observed pixels in $\tilde{\mathbf{D}}$; and $\gamma$ is set to be 0.8 in our experiments. Compared with the $L2$ loss, the Huber loss function $l(\cdot)$ has lower sensitivity to outliers and presents more robust performance at discontinuous areas \cite{huber1992robust}.

Since the $\mathcal{L}_{P}$ term only considers the sparse pixels, we include the $\mathcal{L}_{R}$ term to add constraints on the densely predicted pixels. Inspired by \cite{godard2017unsupervised}, given a pair of stereo images, the left image $\mathbf{I}_l$ can be reconstructed from the right image $\mathbf{I}_r$ based on the estimated disparity image $\hat{\mathbf{D}}$. To make the reconstruction process differentiable, we employ a bilinear sampler in this process. Finally, the $\mathcal{L}_{R}$ term is defined as a combination of a single-scale SSIM term \cite{wang2004image} and an $L1$ norm term:
\begin{equation}
    \begin{split}
        \mathcal{L}_{R}=\frac{1}{N_{\hat{\mathbf{I}}_{l}}}\sum_{\mathbf{p} \in \hat{\mathbf{I}}_{l}} \alpha & \frac{1-\text{SSIM}\left(\mathbf{I}_{l}(\mathbf{p}), \hat{\mathbf{I}}_{l}(\mathbf{p})\right)}{2}\\
        &\hspace{1cm}+(1-\alpha)\left\|\mathbf{I}_{l}(\mathbf{p})-\hat{\mathbf{I}}_{l}(\mathbf{p})\right\|_{1},
    \end{split}
    \label{eq.reconstruction_loss}
\end{equation}
where $\hat{\mathbf{I}}_{l}$ is the reconstructed image from $\mathbf{I}_r$ according to $\hat{\mathbf{D}}$; and $N_{\hat{\mathbf{I}}_{l}}$ is the number of observed pixels in $\hat{\mathbf{I}}_{l}$. Similar to \cite{godard2017unsupervised}, we use a simplified SSIM with a $3\times3$ block filter and set $\alpha=0.85$ in our experiments.

Inspired by \cite{heise2013pm}, we further add the $\mathcal{L}_{S}$ term to smooth the disparity predictions. Since the disparity values at the place where the image pixel intensity changes greatly usually vary significantly, we define the $\mathcal{L}_{S}$ term as weighting the disparity gradients ($\partial\hat{\mathbf{D}}$) with an edge-aware term using the image gradients ($\partial \mathbf{I}_l$):
\begin{equation}
    \mathcal{L}_{S}=\frac{1}{N_{\hat{\mathbf{D}}}} \sum_{\mathbf{p} \in \hat{\mathbf{D}}} |\partial_{x} \hat{\mathbf{D}}(\mathbf{p}) |  e^{-\left\|\partial_{x} \mathbf{I}_{l}(\mathbf{p})\right\|_{1}}+|\partial_{y} \hat{\mathbf{D}}(\mathbf{p})| e^{-\left\|\partial_{y} \mathbf{I}_{l}(\mathbf{p})\right\|_{1}}.
    \label{eq.smooth_loss}
\end{equation}

Furthermore, in traditional stereo matching approaches, the LRDCC is usually performed to refine the estimated disparities, as the occluded areas are only visible in one image. Inspired by the LRDCC, we flip the left and right stereo images in the left-right direction, respectively, as shown in Fig. \ref{fig.framework}. The flipped left and right images are considered as a pair of new right and left images, respectively. This process can augment the training data to improve the stereo matching performance of DCNNs.

\begin{figure}[t]
    \centering
    \includegraphics[width=0.8\linewidth]{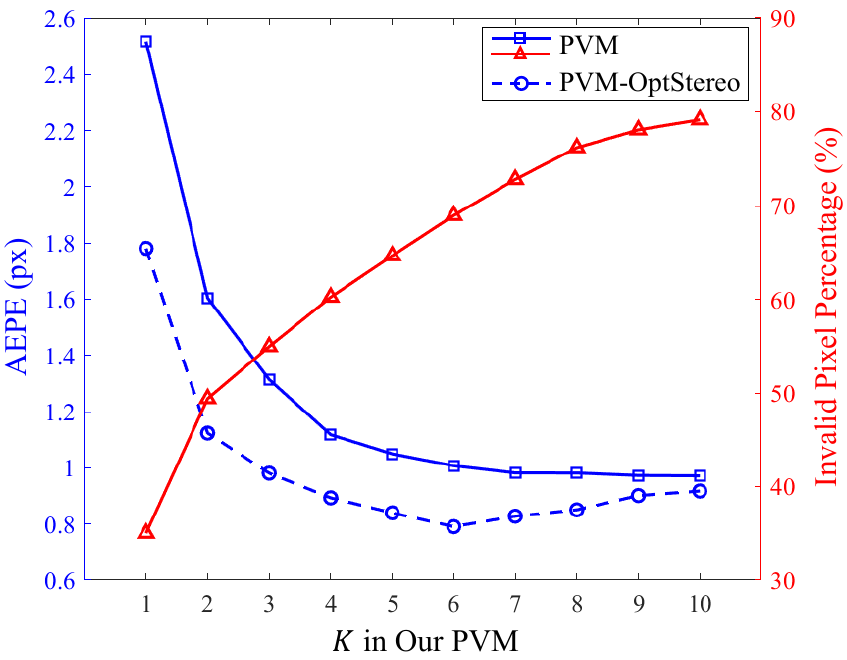}
    \caption{Experimental results of the proposed approach with different variants of our PVM on the HKUST-Drive dataset.}
    \label{fig.ablation_pvm}
\end{figure}

\section{Experimental Results and Discussion}
\label{sec.experiments}

\subsection{Datasets and Implementation Details}
We use three stereo datasets in our experiments:
\begin{itemize}
    \item The KITTI Stereo 2012 benchmark \cite{kitti12}: This dataset contains 194 training stereo image pairs with sparse ground-truth disparity images and another 195 testing stereo image pairs without ground-truth disparity images.
    \item The KITTI Stereo 2015 benchmark \cite{kitti15}: This dataset contains 200 training stereo image pairs with sparse ground-truth disparity images and another 200 testing stereo image pairs without ground-truth disparity images.
    \item Our HKUST-Drive dataset: We publish a large-scale synthetic dataset, named the HKUST-Drive dataset. This dataset is created using the CARLA simulator \cite{dosovitskiy2017carla}. It is collected in six different scenarios under different illumination and weather conditions, \textit{e.g.,} clear, rainy, daytime and sunset. There are a total 11568 pairs of stereo images with corresponding dense sub-pixel disparity ground truth. We split it into a training set (6940 image pairs), a validation set (2314 image pairs) and a testing set (2314 image pairs). Different from the KITTI stereo datasets \cite{kitti12,kitti15}, our HKUST-Drive dataset can effectively evaluate the generalization ability of stereo matching approaches across different weather and illumination conditions.
\end{itemize}

During the training phase, we use the Adam optimizer \cite{kingma2014adam} and adopt an initial learning rate of $10^{-4}$. The model is trained on two NVIDIA GeForce RTX 2080 Ti graphics cards until it converges. We also adopt an existing self-supervision scheme \cite{liu2020learning} to improve the stereo matching performance on the challenging areas, such as the occluded areas.
Moreover, we use two commonly used metrics for evaluation: 1) the average end-point error (AEPE) that measures the average difference between the disparity estimations and ground-truth labels and 2) the percentage of pixels (F1) with absolute disparity error higher than 3 pixels \cite{kitti12}.

In our experiments, we first conduct ablation studies on our HKUST-Drive dataset in Section \ref{sec.ablation} to 1) select the best architecture of our PVM and OptStereo as well as the best hyper-parameters for the loss function; and 2) demonstrate the effectiveness of our data augmentation technique. Then, we denote our best self-supervised implementation as \textit{PVStereo}, and compare it with the state-of-the-art self-supervised approaches on our HKUST-Drive dataset. We also compare our OptStereo with the state-of-the-art supervised approaches. The experimental results are shown in Section \ref{sec.hkust-drive}. After that, we evaluate our PVStereo and OptStereo on the popular KITTI Stereo 2012 and 2015 benchmarks \cite{kitti12,kitti15}, as presented in Section \ref{sec.benchmark}.

\begin{table}[t]
    \centering
    \caption{Experimental Results of the Proposed Approach with Different Variants of Our OptStereo on the HKUST-Drive Dataset. The Adopted Variant is Bolded}
    \begin{tabular}{C{0.5cm}C{3.0cm}C{1.8cm}C{1.8cm}}
        \toprule
        No. & \makecell{Multi-scale                           Cost Volumes} & \makecell{Lookup                 \\Distance $d$} & AEPE (px) \\ \midrule
        (a) & --                                                            & 4                & 1.02          \\ \midrule
        (b) & \cmark                                                        & 1                & 1.16          \\
        (c) & \cmark                                                        & 2                & 0.93          \\ \midrule
        (d) & \cmark                                                        & 4                & \textbf{0.79} \\
        \bottomrule
    \end{tabular}
    \label{tab.ablation_optstereo}
\end{table}

\begin{figure}[t]
    \centering
    \includegraphics[width=0.99\linewidth]{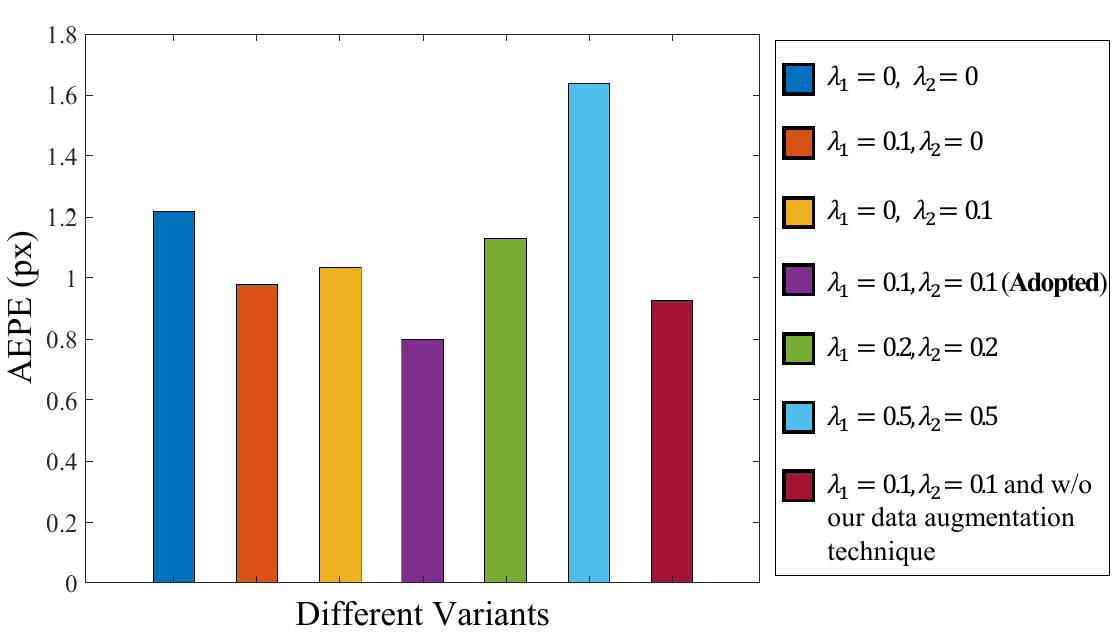}
    \caption{Experimental results of the proposed approach with different variants of our loss function and data augmentation technique on the HKUST-Drive dataset.}
    \label{fig.ablation_loss}
\end{figure}

\begin{figure}[t]
    \centering
    \includegraphics[width=0.9\linewidth]{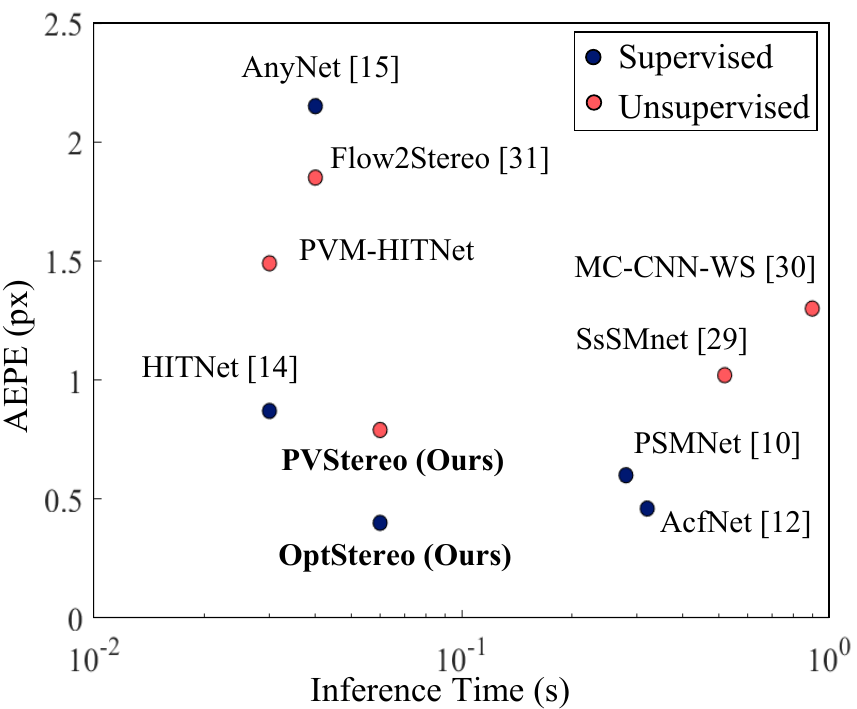}
    \caption{Performance comparison of our OptStereo and PVStereo with state-of-the-art supervised and unsupervised approaches on the HKUST-Drive dataset.}
    \label{fig.hkust-drive-evaluation}
\end{figure}

\subsection{Ablation Study}
\label{sec.ablation}
We first explore the best architecture of our PVM. Fig. \ref{fig.ablation_pvm} compares the performance of our PVM and the corresponding PVM-OptStereo with respect to different $K$. We can observe that with $K$ increasing, the AEPE of the PVM decreases while the invalid pixel percentage increases. Correspondingly, the AEPE of our PVM-OptStereo first decreases but then increases. Therefore, we minimize the trade-off between accuracy and density, and set $K$ to 6 in the rest of our experiments, where the corresponding PVM-OptStereo can achieve the best performance. Additionally, we can also observe that all the variants of our PVM-OptStereo outperform the corresponding PVM. We believe that our designed loss function and proposed data augmentation technique can provide effective fine-tuning for the predicted dense disparity images, thus making them outperform the corresponding semi-dense disparity images used for training.

Table \ref{tab.ablation_optstereo} compares the performance of the proposed approach with different variants of our OptStereo. The comparison between (a) and (d) demonstrates the effectiveness of the adopted multi-scale cost volumes. From (b)--(d), we can observe that with the lookup distance $d$ increasing, the performance of our approach improves. Considering the balance between accuracy and efficiency, we adopt (d) of Table \ref{tab.ablation_optstereo} in the rest of our experiments.

In addition, we test different combinations of $\lambda_{1}$ and $\lambda_{2}$ in the loss function, and some of the experimental results are presented in Fig. \ref{fig.ablation_loss}. We can observe that the proper introduction of $\mathcal{L}_R$ and $\mathcal{L}_S$ can improve the performance effectively. We analyze that $\mathcal{L}_R$ and $\mathcal{L}_S$ can perform effective supervision for the pixels that are invalid in our PVM, and thus, can benefit the overall performance of dense stereo matching. However, $\lambda_{1}$ and $\lambda_{2}$ must be kept low to not overcome the contribution of $\mathcal{L}_P$. Based on Fig. \ref{fig.ablation_loss}, we set $\lambda_{1}=0.1$ and $\lambda_{2}=0.1$ in the rest of our experiments. Furthermore, Fig. \ref{fig.ablation_loss} demonstrates that our data augmentation technique can effectively improve the stereo matching performance. We analyze that the proposed technique can leverage the relationship among stereo images to perform effective training data augmentation, and thus, can benefit the overall performance of dense stereo matching.

\begin{figure*}[t]
    \centering
    \includegraphics[width=0.9\textwidth]{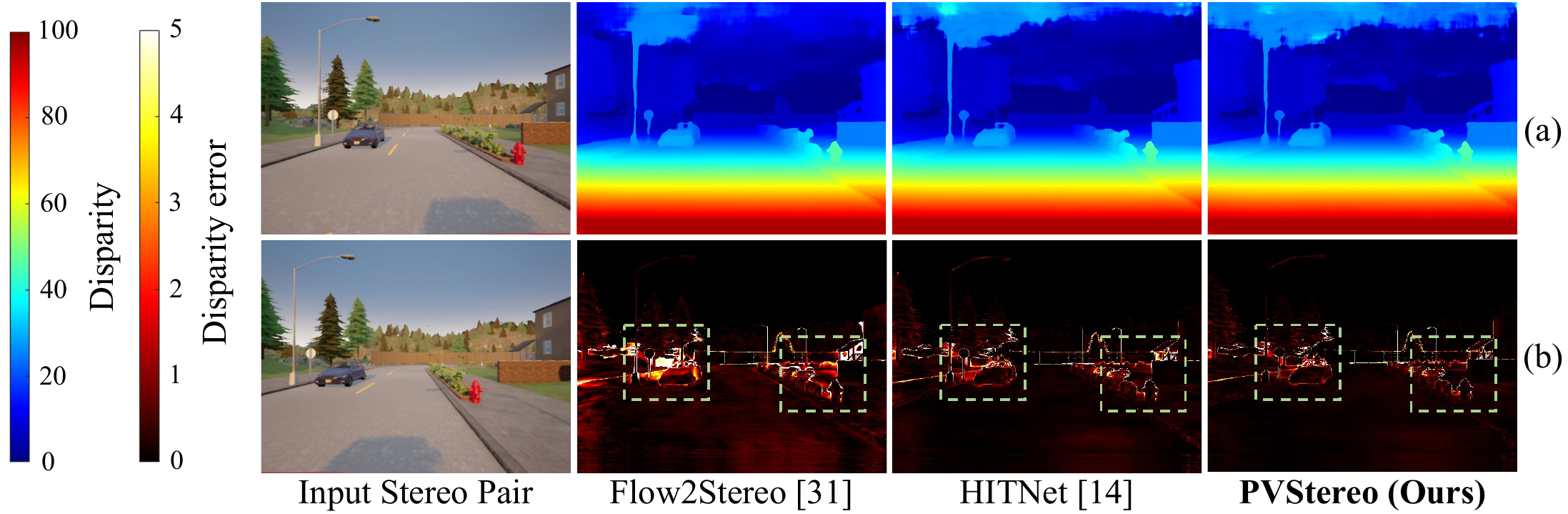}
    \caption{An example on our HKUST-Drive dataset, where rows (a) and (b) show the disparity estimations and the corresponding disparity error maps, respectively. Significantly improved regions are marked with green dashed boxes.}
    \label{fig.hkust-drive}
\end{figure*}

\subsection{Performance Comparison on Our HKUST-Drive Dataset}
\label{sec.hkust-drive}
As previously mentioned, we denote our best self-supervised implementation as \textit{PVStereo}, and compare it with the state-of-the-art self-supervised approaches on our HKUST-Drive dataset. We also train our OptStereo in a supervised manner, and compare it with the state-of-the-art supervised approaches. In addition, we implement our PVM in existing HITNet \cite{tankovich2020hitnet}, and denote it as PVM-HITNet. For the other baseline supervised and unsupervised approaches, we follow the hyperparameter setups as reported in their papers.

The quantitative results are presented in Fig.~\ref{fig.hkust-drive-evaluation}. We can clearly observe that our PVStereo significantly outperforms all other state-of-the-art self-supervised approaches. Moreover, our PVStereo can achieve a great trade-off between accuracy and efficiency for stereo matching. It is also evident that PVM-HITNet can present competitive performance for self-supervised stereo matching, which verifies the effectiveness of our PVM. Furthermore, our PVStereo can even achieve competitive performance compared to supervised approaches such as HITNet \cite{tankovich2020hitnet}, which demonstrates the effectiveness of our proposed self-supervised architecture. Please note that our OptStereo achieves a better performance than state-of-the-art supervised approaches, which verifies the superiority of the proposed DCNN architecture over existing architectures that rely on 3D convolutions or coarse-to-fine paradigms. Fig.~\ref{fig.hkust-drive} illustrates an example of the qualitative results. We can see that our PVStereo can generate more accurate disparity estimations. Since our HKUST-Drive dataset covers different scenarios under different illumination and weather conditions, these results strongly demonstrate the great generalization ability of our OptStereo and PVStereo across different scenarios as well as different weather and illumination conditions.

\begin{table}[t]
    \caption{Evaluation Results on the KITTI Stereo 2012$^{1}$~\cite{kitti12} and KITTI Stereo 2015$^{2}$ \cite{kitti15} Benchmarks. ``Noc'' and ``All'' Represent the F1 (\%) for Non-Occluded Pixels and All Pixels, Respectively. ``S'' Denotes Supervised Approaches, and pSGM~\cite{lee2017memory} is A Traditional Approach. Best Results for Supervised and Unsupervised Approaches are Both Bolded}
    \centering
    \begin{tabular}{L{2.2cm}C{0.2cm}C{0.6cm}C{0.6cm}C{0.6cm}C{0.6cm}C{0.9cm}}
        \toprule
        \multicolumn{1}{l}{\multirow{2}{*}{Approach}} & \multicolumn{1}{c}{\multirow{2}{*}{S}} & \multicolumn{2}{c}{KITTI 2012} & \multicolumn{2}{c}{KITTI 2015} & \multicolumn{1}{c}{\multirow{2}{*}{\makecell{Runtime                                 \\(s)}}} \\ \cmidrule(l){3-4} \cmidrule(l){5-6}
                                                      &                                        & Noc                            & All                            & Noc                                                  & All           &               \\ \midrule
        PSMNet \cite{psmnet}                          & \cmark                                 & 1.49                           & 1.89                           & 2.14                                                 & 2.32          & 0.41          \\
        GwcNet-g \cite{gwcnet}                        & \cmark                                 & 1.37                           & 1.70                           & 1.92                                                 & 2.11          & 0.32          \\
        AcfNet \cite{zhang2020adaptive}               & \cmark                                 & \textbf{1.17}                  & \textbf{1.54}                  & 1.72                                                 & 1.89          & 0.48          \\
        \textbf{OptStereo (Ours)}                     & \cmark                                 & 1.20                           & 1.61                           & \textbf{1.36}                                        & \textbf{1.82} & \textbf{0.10} \\ \midrule
        pSGM \cite{lee2017memory}                     & --                                     & 4.68                           & 6.13                           & 5.17                                                 & 5.97          & 7.77          \\
        Flow2Stereo \cite{flow2stereo}                & --                                     & 4.58                           & 5.11                           & 6.29                                                 & 6.61          & \textbf{0.05} \\
        MC-CNN-WS \cite{tulyakov2017weakly}           & --                                     & 3.02                           & 4.45                           & 4.11                                                 & 4.97          & 1.35          \\
        SsSMnet \cite{zhong2017self}                  & --                                     & 2.30                           & 3.00                           & 3.06                                                 & 3.40          & 0.80          \\
        \textbf{PVStereo (Ours)}                      & --                                     & \textbf{1.98}                  & \textbf{2.47}                  & \textbf{2.69}                                        & \textbf{2.99} & 0.10          \\ \bottomrule
    \end{tabular}
    \label{tab.kitti}
\end{table}

\subsection{Evaluation Results on the KITTI Stereo Benchmarks}
\label{sec.benchmark}
\begin{figure*}[t]
    \centering
    \includegraphics[width=0.99\textwidth]{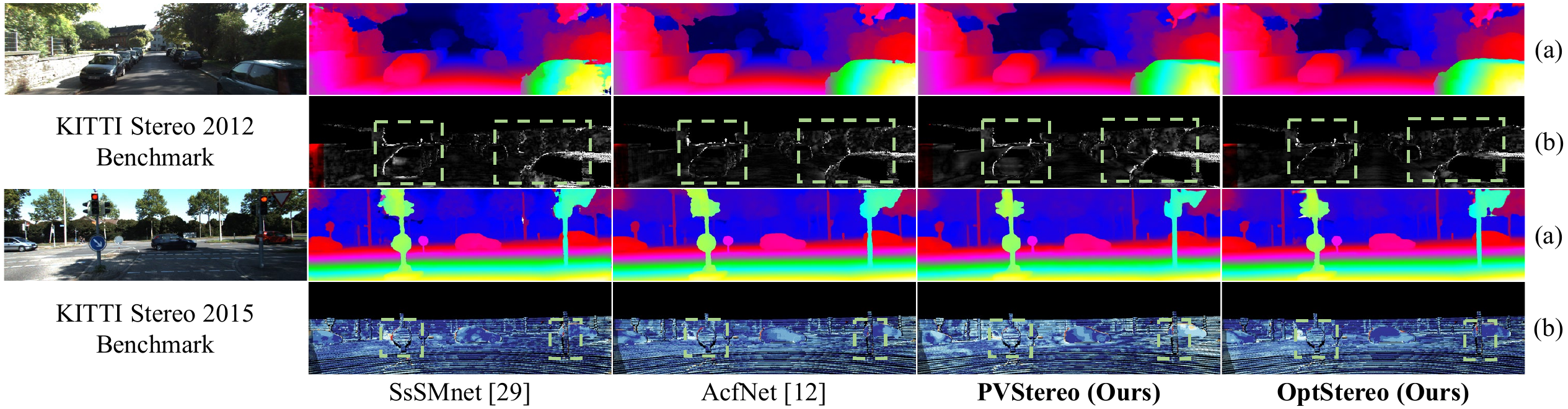}
    \caption{Examples on the KITTI Stereo benchmarks \cite{kitti12,kitti15}, where rows (a) and (b) show the disparity estimations and the corresponding disparity error maps, respectively. Significantly improved regions are highlighted with green dashed boxes.}
    \label{fig.kitti}
\end{figure*}

We submit the results achieved by our OptStereo and PVStereo to KITTI Stereo benchmarks \cite{kitti12,kitti15}, and the quantitative results are presented in Table~\ref{tab.kitti}. It is evident that our OptStereo achieves competitive performance compared to state-of-the-art supervised stereo matching approaches, which demonstrates the effectiveness of the proposed DCNN architecture. Moreover, our PVStereo outperforms all other state-of-the-art self-supervised stereo matching approaches with a great trade-off between accuracy and efficiency, which verifies the superiority of our self-supervised stereo matching approach. Furthermore, Fig.~\ref{fig.kitti} illustrates some examples on the KITTI Stereo benchmarks, where we can see that our OptStereo and PVStereo can yield robust and accurate disparity estimations.

\footnotetext[1]{\url{cvlibs.net/datasets/kitti/eval_stereo_flow.php?benchmark=stereo}}
\footnotetext[2]{\url{cvlibs.net/datasets/kitti/eval_scene_flow.php?benchmark=stereo}}

\section{Conclusions}
\label{sec.conclusions}
This paper presented PVStereo, a novel self-supervised approach for end-to-end stereo matching, which consists of a PVM and a novel DCNN architecture, referred to as OptStereo. Specifically, our OptStereo first builds multi-scale cost volumes, and then adopts a recurrent unit to iteratively update disparity estimations at high resolution, which can not only avoid the error accumulation problem in coarse-to-fine paradigms, but also can achieve a great trade-off between accuracy and efficiency due to its simple but effective architecture. Moreover, our PVM can generate reliable semi-dense disparity images, which can be employed to supervise the training of OptStereo. Furthermore, we publish a large-scale synthetic stereo dataset, named the HKUST-Drive dataset, collected under different illumination and weather conditions for research purposes. Extensive experiments on the popular KITTI Stereo benchmarks and our HKUST-Drive dataset demonstrate the effectiveness and efficiency of our PVStereo, which greatly outperforms all other state-of-the-art self-supervised stereo matching approaches. We believe that our PVStereo can be employed in many robotics applications, such as freespace detection, to improve their performance. It is also promising to employ the proposed architecture in other self-supervised tasks, such as self-supervised optical flow estimation.

\bibliographystyle{IEEEtran}
\bibliography{egbib}

\end{document}